\documentclass[sigconf]{acmart}

\acmConference[ICPC 2022]{The 30th International Conference on Program Comprehension}{May 21–22, 2022}{Pittsburgh, PA, USA}
\usepackage{listings}
\usepackage{tcolorbox}

\copyrightyear{2022}
\acmYear{2022}
\setcopyright{acmcopyright}\acmConference[ICPC '22]{30th International Conference on Program Comprehension}{May 16--17, 2022}{Virtual Event, USA}
\acmBooktitle{30th International Conference on Program Comprehension (ICPC '22), May 16--17, 2022, Virtual Event, USA}
\acmPrice{15.00}

\begin{document}

\title{LAMNER: Code Comment Generation Using Character Language Model and Named Entity Recognition}


\author{Rishab Sharma}
\affiliation{
    \institution{University of British Columbia}
    \country{Canada}
}
\email{rishab.sharma@alumni.ubc.ca}

\author{Fuxiang Chen}
\affiliation{
    \institution{University of British Columbia}
    \country{Canada}
}
\email{fuxiang.chen@ubc.ca}

\author{Fatemeh Fard}
\affiliation{
    \institution{University of British Columbia}
    \country{Canada}
}
\email{fatemeh.fard@ubc.ca}


\newcommand{\todoc}[2]{{\textcolor{#1} {\textbf{[[#2]]}}}}
\newcommand{\todoblue}[1]{\todoc{blue}{\textbf{[[#1]]}}}

\newcommand{\fuxiang}[1]{\todoblue{Fuxiang: #1}}

\begin{abstract}

Code comment generation is the task of generating a high-level natural language description for a given code method/function. 
Although researchers have been studying multiple ways to generate code comments automatically, previous work mainly considers representing a code token in its entirety semantics form only (e.g., a language model is used to learn the semantics of a code token), and additional code properties such as the tree structure of a code are included as an auxiliary input to the model. 
There are two limitations: 1) Learning the code token in its entirety form may not be able to capture information succinctly in source code, and 
2) The code token does not contain additional syntactic information, inherently important in programming languages.

In this paper, we present LAnguage Model and Named Entity Recognition (LAMNER), a code comment generator capable of encoding code constructs effectively and capturing the structural property of a code token. 
A character-level language model is used to learn the semantic representation to encode a code token. For the structural property of a token, a Named Entity Recognition model is trained to learn the different types of code tokens. These representations are then fed into an encoder-decoder architecture to generate code comments.
We evaluate the generated comments from LAMNER and other baselines on a popular Java dataset with four commonly used metrics.
Our results show that LAMNER is effective and improves over the best baseline model in BLEU-1, BLEU-2, BLEU-3, BLEU-4, ROUGE-L, METEOR, and CIDEr by 14.34\%, 18.98\%, 21.55\%, 23.00\%, 10.52\%, 1.44\%, and 25.86\%, respectively.
Additionally, we fused LAMNER's code representation with the baseline models, and the fused models consistently showed improvement over the non-fused models.
The human evaluation further shows that LAMNER produces high-quality code comments.

\end{abstract}

\ccsdesc[500]{Computing methodologies~Artificial intelligence}

\keywords{code comment generation, code summarization, character language model, named entity recognition}




\maketitle

\section{Introduction}


Maintaining the source code documentation is an important software engineering activity. 
It has been reported that software developers are spending more than half of their time trying to understand code during the software maintenance cycle \cite{retref}.
Although program comprehension is the main activity for software developers \cite{ppf}, reading code requires additional mental effort as it is not a natural practice for humans \cite{difnlpl}. 
A well-commented code aids in easier code comprehension and assists in the efficient maintenance of software projects \cite{sridhara}. Despite the importance of well-documented code, the previous study has reported that only a small percentage of the methods in software projects are commented \cite{codernn}. Moreover, comments are mostly absent or outdated as the software evolves \cite{evolve}. 
As a result, researchers have studied on how to generate natural language code comments from a given code method/function automatically \cite{hu1, hu2, iyer, transformer}. 
The code comments aim to explain what the code is doing so that developers do not have to inspect every line of code to infer how it works. This often happens when new developers join a new code repository, or when developers revisit a code repository that has been inactive for some time. 

Since 2016, multiple studies have leveraged the deep learning-based encoder-decoder Neural Machine Translation models (NMT) for comment generation \cite{iyer, hu1, hu11, retref, maclatest, hu2, cocogum, rencos}. 
The encoder in a general NMT model takes a sequence of tokens from a language as input (e.g., English), and the decoder generates the translation of the input sequence into another language (e.g., German) \cite{maclatest}. 
The code comment generation problem can be seen as a translation task between the code (programming language) and the natural language text, which maps an input code snippet to comment in English as output \cite{maclatest}. 
However, programming languages and natural languages have several dissimilar features \cite{difnlpl,gros2020code}. Programming languages are repetitive, mainly due to their syntax, and they can have an infinite number of vocabulary based on the developers' naming choices for identifiers \cite{dls, bigcode}.
Therefore, the NMT techniques used for the translation of natural languages require specific techniques to handle the differences in programming languages \cite{dls}. 

Existing work studied multiple ways to generate code comments automatically. Early neural models for code comment generation \cite{iyer, hu1} used the Long Short Term Memory (LSTM) based encoder-decoder architecture. It mainly considers representing a code token in its entirety semantics form (e.g., a language model is used to learn the semantics of a code token).
Subsequent works \cite{hu1, clair, code2seq, ase} further improved the performance by incorporating additional code properties such as Abstract Syntax Trees (AST) to represent the syntactical structure of code. 
There are two limitations: 1) Learning the code token in its entirety form may not be able to capture information succinctly in source code, e.g., when developers are using camel case or snake case identifiers. Writing camel case or snake case identifiers is a common practice for developers to combine multiple elements together because the identifier names cannot contain spaces \cite{camelidentifier}. For example, writing camel case identifiers is the convention used in Java code. Many existing code comment generation models do not distinguish these identifiers properly. 2) The embedding of the code token does not contain additional structural information, which is inherently important in programming languages. For example, developers need to write the code that conforms to a certain structure for a code to be compiled and executed successfully. 
A code token may be an access modifier, an identifier name, or other code constructs. Code constructs represent the syntactic/structural meaning of each token. The code constructs are reported to be useful in code comprehension and are used in Software Engineering tasks such as bug detection and program repair \cite{cates2021doesIdentifier, newman2020generationIdentifier, allamanis2021Identifier, ir3}.  
We note that AST, which provides a tree structure for code, is another way to represent structural information for code. 
Unlike code constructs, AST cannot encode token-level information.

In this paper, we present LAnguage Model and Named Entity Recognition (LAMNER), 
a code comment generator capable of effectively encoding identifiers and capturing a code token's structural property. To encode a code token that includes a type of identifier, e.g., camel case identifier, snake case identifier, etc., a character-level language model is used to learn the semantic representation. For the structural property of a token, a Named Entity Recognition (NER) model is trained to understand the code constructs.
These representations are then fed into an encoder-decoder architecture to generate code comments. We note here that a code token encoded by our encoder will contain semantic and syntactic information. 
We evaluate the generated comments from LAMNER and other baselines against the ground truth on a popular Java dataset with four commonly used metrics: BLEU, ROGUE-L, METEOR, and CIDEr. 

Our results on code comment generation showed that LAMNER is effective and improves over the best baseline model in BLEU-1, BLEU-2, BLEU-3, BLEU-4, ROUGE-L, METEOR, and CIDEr by 14.34\%, 18.98\%, 21.55\%, 23.00\%, 10.52\%, 1.44\%, and 25.86\%, respectively. Additionally, we fused LAMNER's code representation with baseline models, and the fused models consistently showed improvement over the non-fused models.
We also conducted a qualitative study to evaluate the generated comments from LAMNER. The qualitative results show that the comments generated by LAMNER describe the functionality of the given method correctly, are grammatically correct, and are highly readable by humans.

The primary significance of this work over existing work includes the following:

(1) The learning of the semantics and syntax of a code is effective. Our proposed \textbf{character-based (semantics) embeddings suggested that it can understand Java code better}. Our ablation study on semantics embeddings shows that it does not perform better when our proposed character-based embeddings are replaced and trained with state-of-the-art code embeddings (CodeBERT). We observed degradation in performance. When our proposed character-based embeddings are used on other models (e.g., RENCOS\textsubscript{LAMNER-Embeds}), it improves the performance. 

Separately, the encoding of the syntax within the embeddings is rarely studied. \textbf{Using the syntax embeddings for code summarization is effective, and our experiments showed that the learned syntax embedding helps in generating better summaries.}

(2) LAMNER has high adaptability. Our experiments show that \textbf{the pre-trained embeddings of LAMNER can be combined with the pre-trained embeddings of other approaches to improve their performance, e.g., RENCOS combined with LAMNER forms RECOS\textsubscript{LAMNER-Embeds}.} 
Although exploring other tasks is not the primary focus in this study, the embeddings from LAMNER can be deployed into different tasks, such as bug detection and code clone detection. These tasks may benefit from more information on a programming language's semantics and syntax. We note here that further study is still required for a more comprehensive evaluation.

The contributions of our work are as follows:
\begin{itemize}
  
\item \textbf{A novel code comment generation model, LAMNER, that encodes both the semantic and syntactical information of a code token.} We propose LAMNER that leverages a bidirectional character-level language model and a NER model for learning semantics and the syntactical knowledge of code tokens, respectively.

\item \textbf{Empirical evaluation of the different component contribution in LAMNER.} We perform an ablation study to evaluate different variations of LAMNER, such as generating code comments using the proposed semantic embeddings or the proposed syntactic embeddings.

\item \textbf{Fusing and evaluation of the embeddings learned from LAMNER with baseline models.} We show the adaptability of LAMNER by combining the embeddings learned from LAMNER with baseline models.

\item \textbf{The trained models are open sourced\textcolor{black}{\footnote{\textcolor{black}{https://github.com/fardfh-lab/LAMNER}}}} for replication of the results and usage of the pre-trained embeddings and LAM and NER models in the community.

\end{itemize}

\textcolor{black}{We stress that the main novelty of our work is not to develop a new deep learning architecture but to propose a novel pre-trained embedding that captures both the semantic and syntactic knowledge of the code. Although the role of identifiers and their importance for source code analysis and comprehension is well known, there has been no known technique to represent this combined knowledge as embeddings for code summarization. We show that our proposed pre-trained embedding can enhance the results of other approaches. 
}

\textcolor{black}{The rest of this paper is structured as follows. }
Our approach is described in Section \ref{approach} and we explain the experimental setup in Section \ref{experiments}. Quantitative and qualitative results are presented in Section \ref{results} and \ref{human-evalaution}, respectively. 
We point out the threats to validity in Section \ref{threats} and review the related works in Section \ref{related-work}. Finally, we conclude the paper in Section \ref{conclusion}.

\section{Proposed Approach} \label{approach}

\begin{figure*}[t]
    \includegraphics[width =\linewidth]{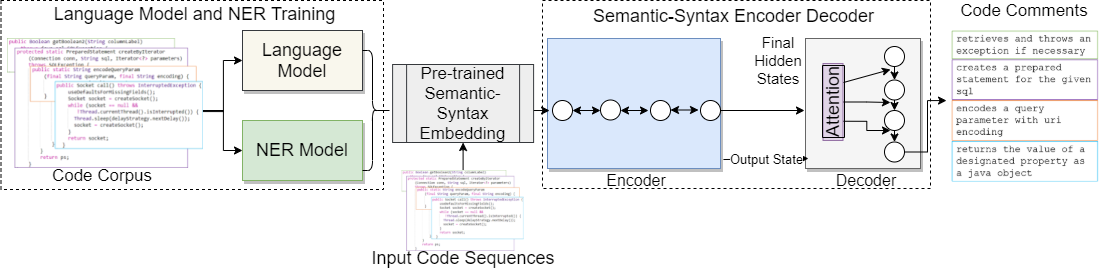}
    \caption{Overview of LAMNER Framework. The Language and NER models on the left box are pre-trained to provide an initial embeddings for the code tokens. The embeddings are then concatenated to serve as the input to the Semantic-Syntax encoder on the right box to generate comments.}
    \label{fig:proposedarch}
\end{figure*}

Figure \ref{fig:proposedarch} illustrates the overview of our proposed approach, LAMNER.
First, a bidirectional character-based language model and a NER model (left box in figure \ref{fig:proposedarch}) are trained separately on a code corpus to generate the input code embeddings for our Semantic-Syntax encoder-decoder architecture (right box in figure \ref{fig:proposedarch}). 
Specifically, the embeddings learned from the language, and the NER models are concatenated for each token. 
These are the extracted embeddings from both models that are concatenated without any pre-processing. For example, if each of the embedding is a vector of size 256, the concatenated embedding will be a vector of size 512. Thus, the concatenated embedding incorporates both the semantic (from the language model) and the syntactic knowledge (from the NER).
We call this the Semantic-Syntax embedding. 
The input to the Semantic-Syntax encoder is the Semantic-Syntax embedding.
The decoder then uses the attention mechanism to decode the input code snippet into code comments.  

The details for the character-based language model, the NER model, and the Semantic-Syntax encoder-decoder are described in Section \ref{sem-code}, \ref{NER} and \ref{semantic-syntax-model}, respectively.

\subsection{Learning Semantics Representation of Code Sequences} \label{sem-code}
We utilize the approach of Akbik et al. \cite{flair} to train and adapt a character-level language model to learn the semantic representation of a code token. It is reported that this language model can capture multiple words within a token better. 
We note here that our aim is not to generate character embeddings but to use a character-level language model to generate embeddings for a code token. We first describe the character-level language model (Section \ref{LM-architecture}) before explaining how it is used to generate embeddings for a code token (Section \ref{tokens-semantic-embedding-extraction}). 

\begin{figure}[b]
    \centering
    \includegraphics[width =\linewidth]{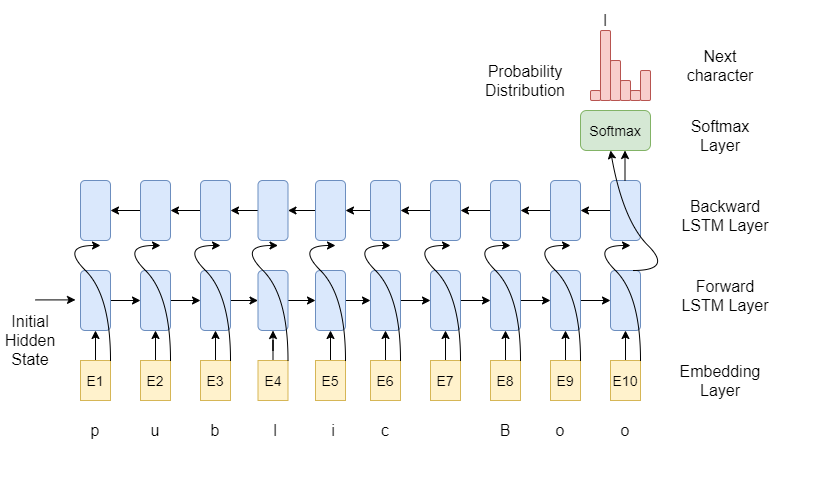}
    \caption{Overview of the bidirectional character-level language model architecture. The input to the model is every character within each line of code, and the model is trained to predict the next character.}
    \label{fig:bidirlm}
\end{figure}

\subsubsection{Character-Level Language Model Architecture} \label{LM-architecture}

Figure \ref{fig:bidirlm} illustrates the architecture of the language model, which employs a single layer bidirectional LSTM (Bi-LSTM) \cite{lstm}. 
Bi-LSTM captures the left, and right context of a character in the sequence \cite{bi-crf} using a forward and a backward language model as shown in Figure \ref{fig:bidirlm}.
The input to each LSTM unit is an embedding of a character initialized randomly. Each LSTM unit processes the embedding to generate the output and the next hidden state for the character. The output of the last LSTM unit is used to select the next output character with maximum likelihood.
In Figure \ref{fig:bidirlm}, the model predicts the character \emph{l} for the given sequence of \emph{public Boo}, which makes last letter of \emph{public Bool} sequence. 

\subsubsection{Extracting Semantic Embedding of Tokens from Language Model} \label{tokens-semantic-embedding-extraction}
As shown in Figure \ref{fig:bidirlm}, the hidden state of each character maintains the contextual information of its previous characters. In the left to right model (Forward LSTM Layer), the \textit{last} character of a token encapsulates the information of all characters of the token based on its left context. Similarly, in the right to left model (Backward LSTM Layer), the \textit{first} character of the token contains the information of the previous token from its right context. To get the embeddings of a token, we concatenate these two embeddings.

\subsection{Learning Syntactical Structure of Code Sequences} \label{NER}

Encoding the syntactical structure of code in models has been shown to improve performance in previous work \cite{hu1, tree2seq, code2vec, code2seq}. 
This section presents our approach to generate the syntactical context of a code token. The difference between the syntactical code structure in the previous work and ours is that the previous work does not consider the syntactical properties within a code token itself. For example, a syntactic code structure -- AST, does not have a direct one-to-one mapping with the code tokens. A code token's syntactical structure is inherently important in programming languages. For example, developers need to write the code to conform to a certain syntactical structure to compile and execute a code successfully.

\begin{figure}[htb]
    \centering
    \includegraphics[width =\linewidth]{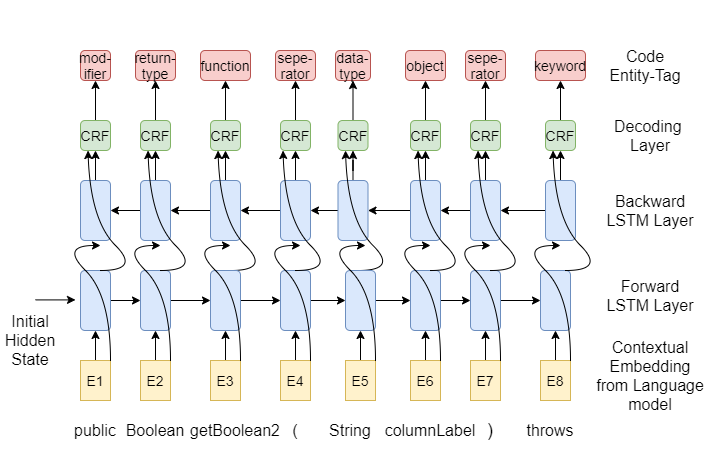}
    \caption{Overview of the NER model to generate the syntactic code embedding for LAMNER. The input to the model is the semantic embeddings of every code token, and they are further fine-tuned to produce the syntactic code embeddings.}
    \label{fig-ner}
\end{figure}

We employ a NER model, shown in Figure \ref{fig-ner}, to generate a code token's syntactical embeddings. It takes the contextual embedding of the input code token from the character language model and uses a bidirectional LSTM sequence tagging with a conditional random field decoding layer \cite{bi-crf} to map each input code token to its syntactic entity type. 
For example, using the NER model for the given code sequence \textit{public Boolean getBoolean2 {...}}, the token \emph{public} is labeled as \textit{modifier}, the token \emph{Boolean} is labeled as \emph{return type}, and the token \emph{getBoolean2} is labeled as \textit{function}.

It has been reported that the embeddings learned from the character-based language model have better performance \cite{flair} than the classical word embeddings such as FastText \cite{fastext}, and Glove \cite{glove} when used as the input to a NER model. Thus, the information to our NER model is the semantic representation learned from the trained character language model described previously in Section \ref{sem-code}.
Within the NER model, the input embeddings are fine-tuned using a bidirectional LSTM before the syntactic types are predicted in the Decoding Layer as shown in Figure \ref{fig-ner}.

In the decoding phase, the CRF cell uses the knowledge from the LSTM output states to predict the correct entity type. 
It is important to note that we do not use the predicted entity types of the NER in LAMNER. Rather, the \textit{extracted syntactical representation of the code tokens} learned from the NER model is used. 
After the NER model is trained, it extracts the syntactic embedding for each code token.

Note that the scope of this work is not to detect these code entities. We believe that parsers can detect the syntactic entities better. In comparison, our work presents a new technique to generate context-sensitive syntactic embeddings of code tokens, which is not available through parsers.

\subsection{Semantic-Syntax Model} 
\label{semantic-syntax-model}
The pre-trained embeddings from the language and NER models are concatenated to represent the semantic-syntax embedding for code tokens in a given code snippet. 
The Semantic-Syntax encoder takes this semantic-syntactic embedding of the \textbf{code tokens} as input and models them together to output the semantic-syntax embeddings of the \textbf{code sequence}. The attentive decoder, which uses this code sequence embeddings, is trained to generate the code comments in natural language. The details of the Semantic-Syntax Encoder and the Semantic-Syntax Decoder are in Section \ref{semantic-syntax-encoder} and \ref{semantic-syntax-decoder}, respectively. 

\subsubsection{Semantic-Syntax Encoder}
\label{semantic-syntax-encoder}

The Semantic-Syntax encoder processes the input using a single layer bidirectional GRU. 
We use GRU as it is reported to have faster training time \cite{rnnfund}, and yet preserve the information for long sequences \cite{gruemp}. 

We denote $E_{t}$ as the semantic-syntax embedding of token $t$,
which is the concatenated pre-trained semantics and syntax embeddings from the language model and the NER model. $E_{t}$ serves as the input to the GRU.
Each GRU state processes the semantic-syntax embedding for the current token, $E_{t}$, in the code sequence and generates a hidden state $h_{t}$.
 
The hidden state of the last token in the sequence, $h_{last}$, contains the sequential information of the complete sequence 
and is formed by concatenating the internal hidden states, $h_{t_{left}}$ and $h_{t_{right}}$, from the left and the right layer.
The hidden state of the last token, $h_{last}$, is then fed into a fully connected linear layer. The formal equation for the fully connected layer, $y_{fc}$, is shown below:
\begin{equation}
    y_{fc} = h_{last}*W_{t} + b\label{eq9}
\end{equation}

where $W_{t}$ and $b$ are the weight matrix and bias values, respectively. The output of the fully connected layer, $y_{fc}$, is then passed into a $tanh$ layer to form the final output $h_{final}$ of the Semantic-Syntax encoder. The formal equation is shown below:

\begin{equation}
  h_{final} = tanh(y_{fc})\label{eq10}
\end{equation}

\subsubsection{Semantic-Syntax Decoder}
\label{semantic-syntax-decoder}
The Semantic-Syntax decoder implements the popular Bahdanau's attention mechanism on a unidirectional GRU, which was reported to have good performance \cite{rnn1}. It uses the Semantic-Syntax encoder's final hidden state to pay attention to the input sequence's important tokens. The attention mechanism prevents the decoder from generating the natural language comments based on a single final context vector. 
Instead, it calculates the attention weights for each input token and pays attention to the tokens with larger attention values during decoding. 
To predict a token, the decoder will look up the semantic embeddings of the comments learned during the training of the NMT model. The decoder will predict the next token until it reaches the maximum sequence length or the end of sentence token $\langle$eos$\rangle$.

\section{Experimental Setup} \label{experiments}

In this section, we describe how we evaluate our approach for its effectiveness, including the research questions, the dataset used, the model training details, as well as the baselines models and the variations of our proposed model for comparison, in Section \ref{research-questions}, \ref{dataset}, \ref{model-training}, and  \ref{baselines}, respectively. 

\subsection{Research Questions}
\label{research-questions}

To evaluate our approach, we seek to address the following
research questions in our experiment:

\textbf{RQ1: (Performance of LAMNER) How does our proposed model perform compared to other baselines?} In this research question, we evaluate the performance of LAMNER with the baseline models using common and popular metrics for comment generation: BLEU-n ($n \in [1,4]$), ROUGE-L, METEOR, and CIDEr.  

\textbf{RQ2: (Contribution of the components in LAMNER, e.g., the semantic component, the syntactic component, etc.) Which parts of our model contribute more to the performance?} 
Rather than using our proposed semantics-syntax embeddings, here, we perform an ablation study to evaluate our model using its different variants.
    
\textbf{RQ3: (Effect of Fusing LAMNER with other models) What is the effect of combining LAMNER with other models for comment generation?}
We investigate the adaptability of LAMNER by integrating its embedding with existing embedding and other models. 

\subsection{Dataset} \label{dataset}

We use the widely used Java dataset for code comment generation collected from popular GitHub repositories by Hu et al. \cite{hu2}. The dataset consists of two parts: the method code and its comment, the first sentence extracted from the Javadoc. In total, there are 69,708 code and comment pairs. The training, validation, and testing set are split distinctly into 8: 1: 1. Following previous work \cite{hu2}, we set the maximum size for code tokens and comment tokens to 300 and 30, respectively. Lengthy code and comments will be truncated up to the maximum size.
 
The statistics for the dataset is provided in 
Table \ref{table: dataset-stat}, under the `\# Records in Dataset' column.

\begin{table}[ht]
\caption{Dataset details for code comment generation}
\label{table: dataset-stat}

\begin{tabular}{ccl}

\hline
\textbf{Split} &\vtop{\hbox{\textbf{\# Records in Dataset}}}\\
\hline

Train & 69,708\\
\hline

Validation & 8,714\\
\hline

Test & 8,714\\
\hline

\end{tabular}
\end{table}

\subsection{Model Training}
\label{model-training}

\subsubsection{Language Models}
In our approach, we train character-based language models. The underlying architecture for the model is the same as described in Section \ref{sem-code}. 

The language model learns the semantic representation of code tokens, and it is trained on the \textit{code} corpus of the training dataset. Over here, we are not interested in the code comments as we only want to learn the code representation. Thus, we exclude all comments (e.g., inline, block, and JavaDoc comments). We 
perform comments removal using the Javalang library as it is reported to have good performance in a previous study \cite{rencos}. 
The language model has a dropout probability of 0.1 applied for regularization purposes.

\subsubsection{Named Entity Recognition Model} \label{ner-training}

We require a labeled dataset to train the NER model where each code token is linked to its corresponding syntactic type. For example, a code token may be associated with an access modifier, operator, or other types.
We use the Javalang parser to obtain a labeled dataset from the training dataset.
The Javalang parser is used in previous studies \cite{ner-pos, ase-ptm} and is reported to have good accuracy in labeling the code tokens into their associated types.
The Javalang parser labels some code tokens in a granular fashion. For example, it groups all types of identifiers (e.g., class name, function name, etc.) into a common label \textit{identifier}, and all types of separators (e.g., end of a method, end of the line, etc.) into another common label \textit{separator}. 

In order to learn the finer nuances of the token type, 
we modify some of the labels that are generated from the Javalang parser.
Specifically, we breakdown the \textit{identifier} and the \textit{separators} types where 
the former is divided into five subtypes: \textit{class}, \textit{function}, \textit{object}, \textit{modifier}, and \textit{return-type}, and  
the latter is divided into three subtypes: \textit{body-start-delimiter}, \textit{body-end-delimiter}, and \textit{end-of-line} (eol).
The breakdown of the identifiers are performed as follow: To identify a \textit{class}, it must conform to the JAVA coding convention i.e., the identifier starts with an upper-case character and it must not contain any braces as suffix. For \textit{function} and \textit{object} identifiers, they must be suffixed by round braces. To distinguish an \textit{object} from an \textit{identifier}, it must have a corresponding \textit{class} identifier with exact naming (\textit{case insensitive}). If there is such a naming match, we categorise it as an \textit{object}, else, we categorise it as a \textit{function}. 
The \textit{return-types} are defined at the start of the function definition, and we identify them directly. Java has a specific set of access modifiers, i.e., static, public, and private, and identifiers that contain them will be labeled as \textit{modifier}. 
For the separators, the token ``\{" is used to indicate the start of a new body -- we labeled it as \textit{body-start-delimiter}. Similarly, we labeled the token ``\}" as \textit{body-end-delimiter} to present the end of a body section. The ``;" token is used to indicate the end of a code line, and we labeled it as \textit{end-of-line} (eol). 

We train the NER model on the training dataset and evaluate its performance on the test dataset. 
The code entity types inferred by the Javalang parser are considered the ground truth labels, whereas the code entities generated by the NER model are the predicted labels. 
We applied dropout with a probability of 0.1 for regularization purposes. 
For evaluating the NER model, we use Precision, Recall, and F1 scores, which are the commonly used metrics to assess NER models in previous studies \cite{flairlib}.
On average, the NER model can achieve 99.41\%, 93.66\%, and 93.89\% for Precision, Recall, and F1 scores, respectively.

As mentioned previously, the goal of this work is not on improving the existing language parser but to generate the syntactical information of each code token.
Specifically, the embeddings learned in the NER model will be used to represent the syntactical information of each code token.
We note here that the Precision, Recall, and F1 scores are used only to evaluate the NER model, the intermediate model used in LAMNER. Other metrics that are used for code comment generation evaluation
are described in Section \ref{metrics}.

Note the difference between Javalang parser and the NER model. The Javalang tool tags each token as a modifier, data type, etc., whereas the NER model is trained to predict the type of the code tokens (i.e. the code constructs, e.g. identifier). If we use the Javalang tool, it can only provide us with the syntactic types of the tokens. These syntactic types are discrete values and we will not be able to incorporate them into our proposed deep learning networks. Therefore, we are unable to use the Javalang parser. By using the NER model, we extract meaningful vector, representations about the type of code tokens which is not possible using the Javalang parser. These embeddings are used in LAMNER.

\subsubsection{Semantic-Syntax Model}

To train the NMT encoder-decoder model, we use the dataset described in section \ref{dataset}. Similar to previous work \cite{codernn, rencos}, the numerical and string values are replaced with the \verb|NUM| and \verb|STR| tokens, respectively. 
The hidden size of the Semantic-Syntax model is set to $512$, and it is trained for $100$ epochs or until the learning rate decays to $1e-7$. The initial learning rate is set to $0.1$, and the batch size is $16$. 
Both encoder and decoder applied a dropout probability of $0.1$.
If the validation loss does not improve after 7 consecutive epochs, the learning rate is decayed by a factor of $0.1$. All experiments are conducted on an NVIDIA Tesla V100 GPU with 32 GB memory. 
\subsection{Baselines and Model Variations} \label{baselines}

\subsubsection{Other Approaches}
We compare our approach with the following baseline models, 
which are commonly used in many comment generation studies \cite{retref, msr-context,rencos, hgnn}. The availability of the models is also another important factor in choosing the baselines.

\textbf{CODE-NN} initializes the code input with one hot vector encoding and uses an LSTM-based encoder-decoder model to generate code comments \cite{iyer}.

\textbf{Hybrid-DRL} employs an actor-critic reinforcement learning approach to generate natural language summaries \cite{ase}. They generate a
hybrid code representation using LSTM and AST, and perform
hybrid attention that follows an actor-critic reinforcement learning architecture.

\textbf{AST-Attend-GRU} uses the code and the Structure-Based Traversal (SBT) representation of AST as input to two separate encoders. The input inside the two encoders is processed and combined to generate the final output \cite{clair}.  

\textbf{TL-CodeSum} uses a double encoder architecture to generate code comments \cite{hu2}. The API knowledge is encoded into embeddings, which are transferred to the model to produce comments.

\textbf{Re2-Com} combines both information retrieval and deep learning-based techniques. A combination of input code, similar code snippets, AST, and exemplar comments generates the code comments.

\textbf{RENCOS} combines the information retrieval techniques with NMT-based models \cite{rencos} . It uses two syntactically and semantically similar code snippets as input. The conditional probability from the two inputs is then fused to generate the final output. 

\textbf{CodeBERT} trains a code-level embeddings model (pre-trained language model) and uses it to perform code summarization through a fine-tuning process \cite{codebert}.

{\textbf{LAMNER\textsubscript{CodeBERT-Embeds}} leverages the same architecture used in LAMNER. However, instead of the LAM and NER embeddings, it employs the embedding extracted from CodeBERT to initialize its vocabulary. \cite{codebert}.}

We use the official implementation available on the authors' GitHub repositories for all of these works. For the Seq2Seq model, its official implementation is from OpenNMT \footnote{https://opennmt.net/} \cite{open-nmt}.

\subsubsection{Variations of Our Model}
We consider four variations of our model as described below. 

\textbf{LAMNER\textsubscript{LAM}} This model uses only the semantic code embeddings learned from the character-level language model.

\textbf{LAMNER\textsubscript{NER}} This model uses only the syntactic code embedding learned from the NER model.

\textbf{LAMNER\textsubscript{Static}}

In this model, the Semantic-Syntax code embeddings are kept static and are not further fine-tuned during training. This model shows the performance of the pre-trained embeddings.

\textbf{LAMNER} This model uses the concatenated semantic and syntactic code embeddings as the input to the Semantic-Syntax encoder, as shown in Figure \ref{fig:proposedarch}. The purpose of this model is to evaluate the effectiveness of combining the two embeddings for code comment generation. Note that in LAMNER, the model is initialized with the Semantic-Syntax code embeddings, and the code embeddings are further fine-tuned during training. 
The fine-tuning of code embeddings means that once the embeddings are extracted and concatenated from the language model and the NER model, they are used to initialize the vocabulary. The vocabulary matrix parameters that are initialized with the LAMNER embeddings are then further learnt with the other model parameters.

\subsection{Evaluation Metrics}\label{metrics}

Similar to previous works \cite{ase, rencos}, we evaluate the performance of our model and the baseline models using the following metrics: \textbf{BLEU} \cite{bleu}, \textbf{ROUGE-L} \cite{rouge}, \textbf{METEOR} \cite{meteor}, and \textbf{CIDER} \cite{cider}.

\textbf{BLEU} measures the n-gram ($n \in [1,4]$) geometric precision $p_n$ between the generated comment (C) and the ground truth (G) \cite{bleu}. 

\textbf{ROUGE} is a recall-based metric that computes the number of tokens from (G) that appears in (C) \cite{rouge}. ROUGE-L finds the F-score of the longest common subsequence (LCS) between the two sentences X and Y with length $m$ and $n$, respectively \cite{rouge}. 
\textbf{METEOR} calculates the semantic score using an alignment approach, in which a unigram in a sentence is mapped to zero or one unigram in another sentence in a phased manner \cite{meteor}. 
\textbf{CIDEr} rates the correctness of the comments \cite{rencos, cocogum}. It performs Term Frequency-Inverse Document Frequency (TF-IDF) weighting for each token and uses cosine similarity between the Reference and candidate sentences \cite{cider}. 

For all the metrics, a higher value is considered a better score.

\section{Results} \label{results}

In this section, we present the results for our research questions (Section \ref{research-questions}).

\subsection{RQ1: (Performance of LAMNER)} \label{rq1}
Table \ref{table:results} shows the results of the baseline models and all the variations of our proposed approach. 
The first column shows the different models used in the evaluation, and columns two to five show the BLEU scores. ROUGE-L, METEOR, and CIDEr scores are shown in columns six, seven, and eight, respectively. 
For all the baselines, we used the best hyperparameters and settings mentioned by the authors of the models to have a fair comparison.

\begin{table*}[ht]
  \caption{Evaluation of various baseline models and our proposed model, LAMNER. LAMNER has the best performance in BLEU\{1-4\}, ROUGE-L, METEOR and CIDEr. Our model variants have also consistently achieve better performance than the baseline models.}
  \label{table:results}
  \begin{center}
  \begin{tabular}{ccccccccl}
    \toprule
   Model & BLEU-1(\%) & BLEU-2(\%) & BLEU-3(\%) & BLEU-4(\%)  & ROUGE-L(\%) & METEOR(\%) & CIDEr\\
    \midrule

    CODE-NN &23.90  &12.80  &8.60 &6.30 &28.90 &9.10 &0.98\\
    
    AST-Attend-GRU &22.00  &10.05  &5.06  &2.79  &24.92 &8.82 &0.30\\
    CodeBERT  &24.73  &18.35  &15.06  &13.16 &34.46 &17.82 &2.01\\
    TL-CodeSum &29.90  &21.30  &18.10  &16.10  &33.20 &13.70  &1.66\\
    Hybrid-DRL &32.40  & 22.60  &16.30  &13.30  &36.50 &13.50 &1.66\\
    
    Re2-Com &33.19  &24.17  &19.63  &17.07  &41.03 &15.70 &1.56\\
    RENCOS &38.32  &33.33  &30.23  &27.91  &41.07 &24.98 &2.50\\
    LAMNER\textsubscript{CodeBERT-Embeds}  &44.35  &35.19  &30.91  &28.48 &47.64 &21.59 &2.63\\
    \hline
    LAMNER\textsubscript{Static}  &49.57  &40.86  &36.68  &34.21 &51.42 &24.52 &3.20\\
    LAMNER\textsubscript{LAM}  &50.11  &41.09  &36.63 &33.98  &52.41 &25.00 &3.23\\
    LAMNER\textsubscript{NER}  &50.54  &41.68  &37.36  &34.83  &52.64 &25.30 &3.27\\
    LAMNER  &\textbf{50.71}  &\textbf{41.87}  &\textbf{37.57}  &\textbf{35.03}  &\textbf{52.65} &\textbf{25.34} &\textbf{3.31}\\
    
     \bottomrule
  \end{tabular}
  \end{center}
\end{table*}

\begin{figure}[htb]
    \centering
    \includegraphics[width =\linewidth]{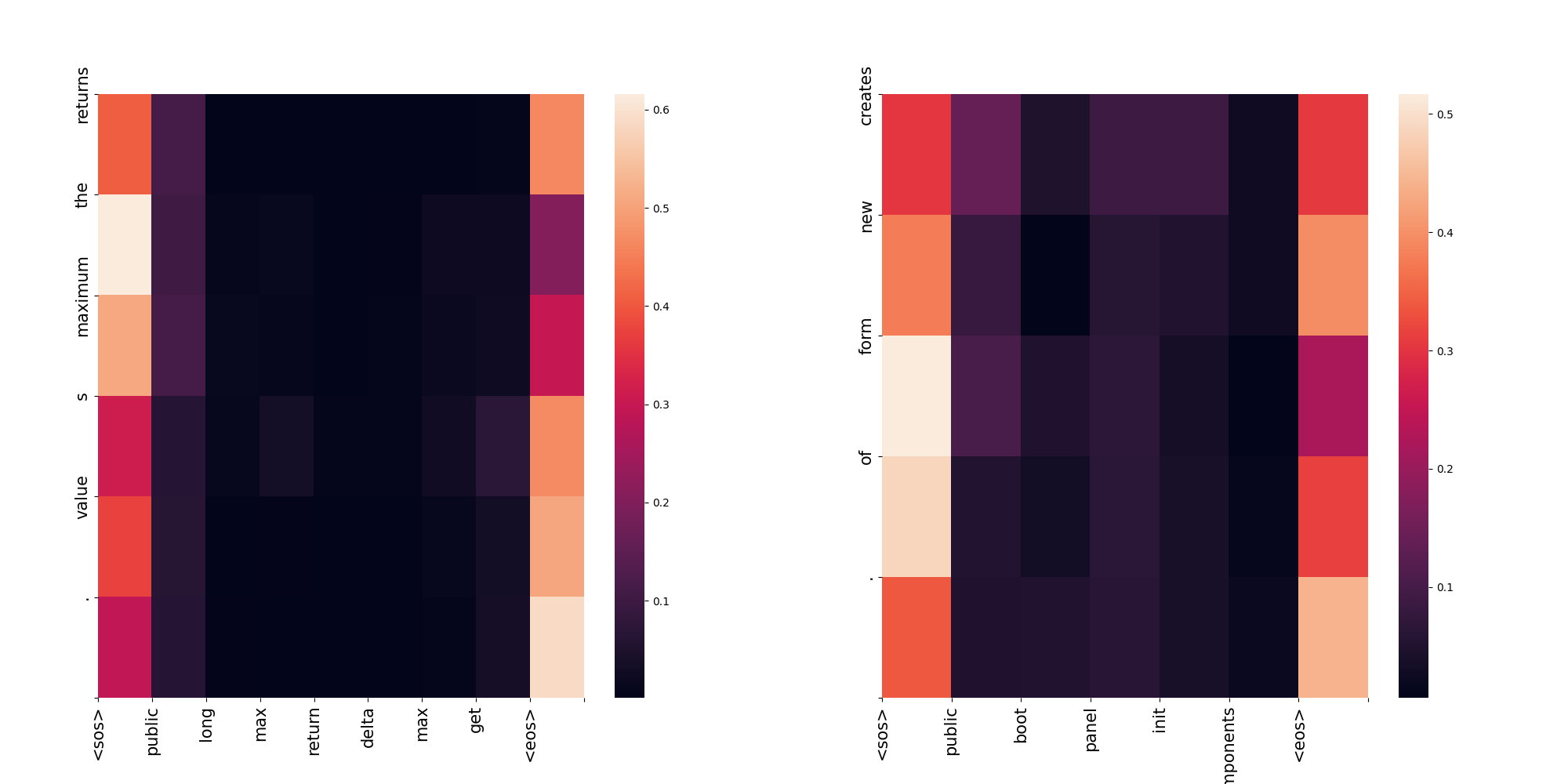}
    \caption{Attention behavior of LAMNER\textsubscript{CodeBERT-Embeds}. Only a few tokens have focused attention. }
    \label{bert-att}
\end{figure}
Among the baselines (rows one to eight), {LAMNER\textsubscript{CodeBERT-Embeds}} has the best scores for BLEU\{1-4\}, ROUGE-L and CIDEr, while RENCOS has the best score for METEOR. 
Our proposed model, LAMNER (last row), has achieve the 
highest scores in BLEU-1, BLEU-2, BLEU-3, BLEU-4, ROUGE-L, METEOR and CIDEr. 
It improves over the best baseline model in BLEU-1, BLEU-2, BLEU-3, BLEU-4, ROUGE-L, METEOR and CIDEr by 14.34\%, 18.98\%, 21.55\%, 23.00\%, 10.52\%, 1.44\%, 25.86\%, respectively.
Interestingly, the results of CodeBERT with the recommended setting of training for 3 epochs are very low.

\begin{tcolorbox}
{Summary of RQ1 results:} \textit{Our proposed approach, LAMNER, is effective and achieve the best score among all the baseline models.}
\end{tcolorbox}

\subsection{RQ2: (Contribution of the components in LAMNER)} \label{rq2}

The results for the variations of our proposed model, LAMNER, are shown in rows nine to eleven in Table \ref{table:results}. 
LAMNER\textsubscript{NER} has higher scores than LAMNER\textsubscript{LAM} for all the evaluation metrics, and both LAMNER\textsubscript{NER} and LAMNER\textsubscript{LAM} have better scores than the baseline models in all the metrics. When compared with LAMNER, which is our proposed model learned from the combination of both the embeddings of NER and LAM, LAMNER improves in all the metrics. 
These results show that our proposed approach is effective, and the learned syntactic and semantic information can improve the results.
 
Even when we use the embeddings as static vectors without further training, i.e., the LAMNER\textsubscript{Static}, the results showed improvement in all the baseline models, except for RENCOS -- there is a slight drop in the METEOR score (1.84\%). Even though there is a small decrease in the METEOR score, we note that all the other metrics have improvements over the baseline models.
When comparing LAMNER\textsubscript{Static} with both NER and LAM, the latter has better performance in the majority of the metrics, except for BLEU-3 and BLEU-4 in LAM.

\begin{tcolorbox}
{Summary of RQ2 results:} \textit{The syntactic embeddings (i.e., NER) has more contribution to the model's performance. Both LAM embeddings and NER embeddings learn meaningful information about the code, and their combination improves the results further.}
\end{tcolorbox}

\begin{figure}[htb]
    \centering
    \includegraphics[width =\linewidth]{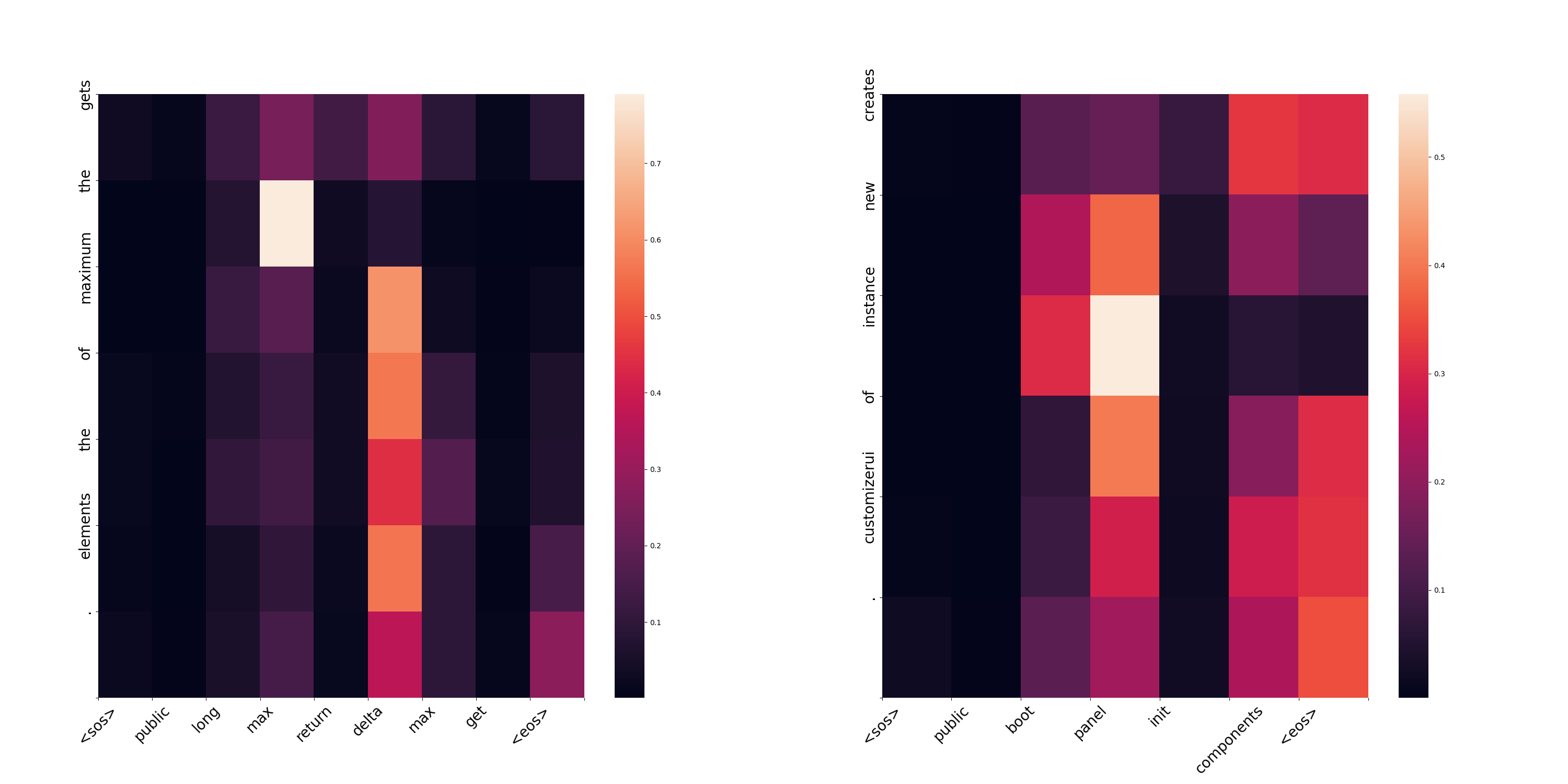}
    \caption{Attention behavior of LAMNER on code samples. The attention is distributed among more tokens.}
    \label{lamner-att}
\end{figure}

\subsection{RQ3: (Effect of Fusing LAMNER with other models)}

To analyze the adaptability of LAMNER, we combine it with other models. 
We integrate LAMNER with existing works in two different ways i) Combining with existing embeddings and ii) Using the pre-trained embeddings of LAMNER with existing models.
For the first approach, we combine the embedding generated from CodeBERT, concatenate it with the LAMNER embeddings (i.e., LAM and NER embeddings), and then use it in the NMT model to generate comments. The model is shown as LAMNER\textsubscript{CodeBERT-Embeds+LAMNER-Embeds}
in Table \ref{table:results-lm-ner}.
Combining LAMNER embedding with CodeBERT embedding improves the LAMNER\textsubscript{CodeBert-Embeds} results by 7.26\% - 10.02\% on BLEU scores, and 3.71\% and 7.5\% for ROUGE-L and METEOR, respectively (CIDEr remains the same).

Additionally, we choose RENCOS, which is the best model among the baselines, and incorporate the LAM and NER embeddings (i.e., LAMNER embedding) into the architecture of RENCOS. In these settings, we initialize each token in the vocabulary of RENCOS with LAMNER embedding. 
This is denoted as RENCOS\textsubscript{LAMNER-Embeds} in Table \ref{table:results-lm-ner}.
Initializing the pre-trained embeddings of LAMNER in RENCOS improves its scores by 6\% to 11\% for all the evaluation metrics.
Here, LAMNER embeddings provide an initial warm-up state for RENCOS, which is then further fine-tuned during the training. 

These results confirm that the embedding learned by the LAMNER model using the character-based language model and NER model can be adapted with other models or combined with different embeddings and are not specific to LAMNER only.

\begin{table*}[ht]
  \caption{(Effect of fusing LAMNER with other
models. For simplicity of comparison, the LAMNER\textsubscript{CodeBERT-Embeds} and RENCOS results from Table \ref{table:results} are repeated here.}
  \label{table:results-lm-ner}
  \begin{center}
  \begin{tabular}{ccccccccl}
    \toprule
   Model & BLEU-1(\%) & BLEU-2(\%) & BLEU-3(\%) & BLEU-4(\%)  & ROUGE-L(\%) & METEOR(\%) & CIDEr\\
    \midrule
    LAMNER\textsubscript{CodeBERT-Embeds}  &44.35  &35.19  &30.91  &28.48 &47.64 &21.59 &2.63\\
    LAMNER\textsubscript{CodeBERT-Embeds+LAMNER-Embeds}  &47.57  &38.37  &34.01  &31.46 &49.41 &23.21 &2.63\\
    \hline
    
    RENCOS &38.32  &33.33  &30.23  &27.91  &41.07 &24.98 &2.50\\
    RENCOS\textsubscript{LAMNER-Embeds}  &42.60  &37.16  &33.66  &30.96 &43.59 &26.89 &2.74\\
    
     \bottomrule
  \end{tabular}
  \end{center}
\end{table*}

\begin{tcolorbox}
{Summary of RQ3 results:} \textit{LAMNER embeddings can be combined with existing works for code comment generation. 
Our proposed Character-level Language models and NER models are effective in creating comments. They further improve the results of other models when used as initialization embeddings or combined with other embeddings. }
\end{tcolorbox}

\section{Example Attention Maps}

Attention plays an important role in the quality of the generated comments \cite{maclatest}. The higher the value of attention, the more important a particular token is during the predictions. Previous works \cite{hu2, clair} used attention behavior to understand the prediction behavior of their models. Figures \ref{bert-att} and \ref{lamner-att} show the attention behavior of the LAMNER\textsubscript{CodeBERT-Embeds} and LAMNER models on two code samples.


For \emph{Sample 1: public long max() \{
return deltaMax.get();\}} ``gets the maximum of the elements ." and ``return the maximums value ." are the comments generated by LAMNER and LAMNER\textsubscript{CodeBERT-Embeds} respectively.
In a similar fashion, for \emph{Sample 2: public BootPanel() \{ initComponents();\}} are ``creates new instance of customizerui ." and ``creates new form of .", are the comments generated by LAMNER and LAMNER\textsubscript{CodeBERT-Embeds} respectively.

We further compare the generated comments with the attention mechanism in LAMNER and LAMNER\textsubscript{CodeBERT-Embeds} in figure \ref{bert-att} and \ref{lamner-att}. We observed that LAMNER has attention distributed to more tokens as compared to LAMNER\textsubscript{CodeBERT-Embeds}, which focused its attention only on a few tokens. This distributive attention allows LAMNER to capture more context of a source code. This can help the model generate a more cohesive latent representation, useful for generating informative code comments. Moreover, the attention mechanism in LAMNER correctly focuses on the more prominent tokens such as `max' and `panel,' which can help determine the exact behavior of the given code samples. We observe that this behavior is missing in the LAMNER\textsubscript{CodeBERT-Embeds} model -- it mostly focuses on other tokens near the start and end of the code sample. We believe that this could be the reason why the code comments generated by the LAMNER\textsubscript{CodeBERT-Embeds} model are not as coherent as LAMNER.

\section{Human Evaluation} \label{human-evalaution}

Automatic metrics are extensively used in machine translation to draw a quantitative comparison among the code summarization models. The models with more overlapping tokens between the references and predictions receive higher scores. However, there can be issues with this evaluation, such as the texts can have the same meaning without using common tokens; thus, despite being a correct prediction, a semantically similar code summary without any overlapping keyword would have zero score. 
This makes it difficult to comprehend the effectiveness of different models \cite{bleurt, bleu-problem}. 

Therefore, we further conducted a qualitative analysis. 
We randomly select 100 generated summaries (for each model) along with their original code, following similar approach to prior research ~\cite{liu2019text, grusky2018newsroom, hu11, iyer, rencos}. Amazon Mechanical Turk (MTurk) workers were hired to rate the quality of the generated summaries, using a rating system where 1 is the worst and 5 is the best score. The MTurkers rated the summary voluntarily, and for each rated comment, the MTurkers are given a compensation of one cent. We used three common criteria to evaluate the summarization quality \cite{liu2019text}: \\
\textbf{Informativeness (I)} How well does the summary capture the key points of the code? \\
\textbf{Relevance (R)} Are the details provided in summary consistent with details in the code? \\
\textbf{Fluency (F)} Are the summaries well-written and grammatically correct? \\ 

Two different workers were required to rate each summary between one and five, where one is the worst and five is the best \cite{iyer, hgnn, tsedrl}.
The MTurkers are shown an example with explanations for all the criteria before starting to rate.
We also ask the MTurkers for their programming coding experience and if they understand the generated summaries and code.
To reduce the bias, the name of the models and the reference comments are not shown to the evaluators. 
Table \ref{table: human} shows the average scores of the models rated by the MTurkers on the four metrics. 
The predictions from \textit{LAMNER} are consistently rated better than RENCOS on all three metrics.

\begin{table}[ht]
\begin{center}

\caption{Results of human evaluation}
\label{table: human}
\begin{tabular}{cccccl}
\hline
Model & Informative & Relevance & Fluency \\
\hline
LAMNER       & 4.13 & 4.18 & 4.13       \\
\hline
RENCOS   & 4.07 & 4.07 &  4.06     \\

\hline

\end{tabular}
\end{center}
\end{table}

\section{Threats to validity} \label{threats}

\textbf{Internal Validity:}
Internal threats in our work relate to the errors in building our models and the replication of the baseline models, as these are the internal factors that might have impacted the results. 
We have cross-checked the implementation of our model for its correctness and will open-source the implementations for easier replication of the results. 
For all the baseline models, we used the official code provided by the authors. 
While training the baseline models, if we encountered any errors or had any doubts, we consulted the authors directly through raising GitHub issues. 
An important difficulty of training the models is preparing the datasets in the required format. 
Various baseline models required separate pre-processing such as generating AST, using third-party libraries such as py-Lucene \footnote{\url{https://lucene.apache.org/pylucene/}}, and extracting the API knowledge. 
For these steps, we followed the instructions provided by the authors closely to ensure the correctness of the required input.

Another threat to validity could be the dataset that we have built for the NER model. We used the Javalang parser to label the dataset. 
This parser reliably tags the Java methods and has been used in several software engineering studies \cite{rencos, ase-ptm}. 
We have broken down the \textit{identifiers} into four sub-categories using a rule-based approach. We take careful steps in breaking down the \textit{identifiers} -- we randomly checked the sub-categories and did not find any misclassification.

\textbf{External Validity:}
In our study, threats to external validity relate to the generalizability of the results \cite{hu11}.
We used an external and extensively used dataset in our work. 
We only applied our model to the Java programming language. Although we hypothesize that the embeddings used in our work are beneficial to other programming languages, more studies are required. 

\textbf{Construct Validity:}
In our work, the automatic evaluation metrics might affect the validity of the results. 
We used four different automatic machine translation metrics (BLEU\{1-4\}, ROUGE-L, METEOR, and CIDEr) to reduce the bias of a particular metric. 
These metrics are frequently used in natural language processing and other related and similar studies in the software engineering domain \cite{codebert, code2seq, rencos, ase, hu11, retref}.

\textbf{Conclusion Validity:}
The conclusion validity refers to the researchers' bias or the bias found in the statistical analysis that can lead to weak results \cite{ampatzoglou2019ConclusionValidity}.
The nature of this study does not depend on the researchers' bias. The conclusion threat can be related to the reliability of the measurements obtained by the automatic metrics. 
To mitigate this threat, we used different evaluation metrics (BLEU\{1-4\}, ROUGE-L, METEOR, and CIDEr) and applied the same calculation to evaluate all the models. 
We note that the automatic metrics cannot fully quantify the quality of the generated comments.
In mitigation, we conducted human studies to compare the results from different models based on the developers' perspectives. 
A potential threat could be related to the conclusions obtained by the human study. To reduce this threat, we anonymized the models, and the reference comments were not shown to the evaluators. Additionally, each generated comment is rated by three evaluators for consistency purposes.

\section{Related Works} \label{related-work}
This section summarizes the related works on code embedding and code comment generation. 
\subsection{Code Embedding}
The research on source code embedding is wide and has many applications \cite{zchen}. 
Hindle et al. \cite{devanbu} used n-grams to build a statistical language model and showed that programming languages are like natural languages: they are repetitive and predictable. 
Bielik et al. proposed a statistical model that applies to character-level language modeling for program synthesis \cite{bielik}.
Recent embeddings for code tokens are based on deep neural networks and are mostly generated with the word2vec model \cite{mikolov2013word2vec} for C/C++, JavaScript, and Java tokens. 
These embeddings are used for program repair \cite{cv3}, software vulnerability detection \cite{cv2}, type prediction \cite{type-pred} and bug detection \cite{cv5}. 
Other embedding techniques from the natural language like FastText \cite{fastext} is used in another work to provide pretrained embeddings for six programming languages \cite{code-fasttext}.
Wang et al. \cite{trace-embed} proposed a technique for program representations from a mixture of symbolic and concrete execution traces.
A modified Graph Neural Network called Graph Interval Neural Network is used to provide a semantic representation of programs \cite{ginn}.
Kanade et al. \cite{cubert} trained BERT \cite{bert} to generate the contextual embeddings and showed their effectiveness on variable misuse classification and variable misuse localization and repair. 
Karampatsis and Sutton use ELMO \cite{scelmo} for bug prediction. 
Alon et al. generate the embeddings of Java methods using AST \cite{code2vec}. 

Lu et al. propose a new embedding in a hyperbolic space that uses function call graphs \cite{hyper}. 
Other works propose neural probabilistic language model for code embedding \cite{fse}, function embeddings for repairing variable misuse in Python based on AST \cite{semcode}, embedding of methods for code clone detection using AST node type and node contents \cite{recursiveast}, embeddings based on tree or graph based approaches \cite{ltrgraph, g1}, embeddings to learn representation of edits \cite{g2}, and embeddings for program repairs \cite{g3}.
Chen at al. \cite{zchen} provides a comprehensive review of the embeddings for source code.
The most similar works are \cite{hu1, nips} and they use word-level encoding for code comments. 
Although code embeddings are widely used in many applications, the existing code embeddings have the limitation of not being able to detect code constructs such as those in camel case and snake case succinctly, and the code token embeddings are not able to capture the structural property in programming languages.
To tackle the limitation, our work proposes a novel Symentatic-Syntax encoder-decoder model, LAMNER. Our experiments showed that LAMNER is effective and has improved performance overall in the baseline models.

\subsection{Code Comment Generation}
Software engineering researchers have proposed multiple techniques to improve automatic code comment generation. 
Initial efforts were made using the information retrieval, template-based, and topic modeling approach. 
Haiduc et al. \cite{ir1} used text retrieval techniques such as Vector Space Model and Latent Semantic Indexing to generate code comments. 
A topic modeling approach was followed by Eddy et al. \cite{eddy} to draw a comparison between their work and of the approach used in \cite{ir1}.
Moreno et al. \cite{mor} used a template-based approach to generate the comments for methods and classes automatically. 
Sridhara et al. \cite{sridhara} introduced Software Word Usage Model (SWUM)  to capture code tokens' occurrences to generate comments. 
Later, Iyer et al. \cite{iyer} presented a neural network for code comment generation. They were the first to use an attention-based LSTM neural translation model for comment generation. 
Hu et al. \cite{hu1, hu11} introduced a model that uses AST. They proposed a modified depth-first search-based traversal algorithm, namely Structure-Based Traversal, to flatten the AST. 
Shahbazi et al. \cite{shahbazi} and Hu et al. \cite{hu2} leveraged API available in the source code to generate summaries. The former leveraged the text content of API's whereas \cite{hu2} used the API names in their respective approaches.
Alon et al. \cite{code2seq} consider all pairwise paths between leaf nodes of AST and concatenate the representation of this path with each leaf node's token representation.
LeClair et al. \cite{clair} presented a dual encoder model that combines the code sequence and AST representation of code.
Liang et al. \cite{codernn} made changes to GRU architecture to enable encapsulating the source code's structural information within itself.
Wan et al. \cite{ase} employed actor-critic reinforcement learning and Tree-RNN to generate comments. 
Yao et al. \cite{coacor} modeled the relationship between the annotated code and the retrieved code using a reinforcement learning framework and used it to generate the natural language code annotations. 
Wei et al. \cite{nips} proposed a dual framework that leverages dual training of comment generation and code generation as an individual model. 
Leclair et al. \cite{maclatest} improved the quality of generated comments by employing a Graph Neural Network model with AST and source code.
A recent approach combines the techniques available in information retrieval to train an NMT model \cite{rencos}. Two similar code snippets are retrieved from the test data and used as input along with the test sequence during testing. 
Similarly, Wei et al. \cite{retref} input a similar code snippet, AST, code sequence, and an exemplar comment to generate better comments. 
In another work, Li et al. \cite{editsum} leveraged a retrieval-based technique to generate the correct keywords within the code comments. It first creates a summary template -- a similar summary retrieved from the training corpus and modified to keep only the important keyword related to the code. This template summary provides a repetitive structure of the code comment which can be edited to replace important keywords from the code.

Liu et al. \cite{jit} and Panthaplackel et al. \cite{jit2} proposed a comment update technique that learns from code-comment changes and generates new comments. 
Wang et al. \cite{cocogum} use code token, AST, intra-class context from the class name, and Unified Modeling Language diagrams to generate comments. 
Haque et al. \cite{msr-context} use full code context file to generate comments for methods. 
More recently, researchers have also become interested in employing a pretrained language model \cite{cubert}. Feng et al. \cite{codebert} trained a multilingual transformer-based language model on six languages and tested the model for code comment generation.
\textit{The previous research used different techniques to represent code. 
This work introduced a novel technique to capture the semantic-syntax information that is inherently important in the programming language.}

\section{Conclusion} \label{conclusion}

This paper presents a novel code comment generation model, LAMNER, which uses semantic-syntax embeddings that encodes a code token's semantic and syntactic structure. 
LAMNER can be combined with other code comment generation models to improve model performance.
The evaluation on BLEU\{1-4\}, ROUGE-L, METEOR, and CIDER metrics confirm that LAMNER achieves state-of-the-art performance. 
We relate this result to the pre-trained embeddings introduced and their ability to extract unseen code sequences' semantic and syntactic representation. 
This result is also supported through human evaluation. The human evaluation also suggests that the comments from LAMNER are fluent and are grammatically correct.  
Several studies conducted show the importance of both the embeddings for comment generation.
In the future, we plan to apply LAMNER to other programming languages and on different tasks such as bug prediction.

\begin{acks}
This research is support by a grant from Natural Sciences and Engineering Research Council of Canada RGPIN-2019-05175.
\end{acks}

\break
\balance
\bibliographystyle{ACM-Reference-Format}
\bibliography{main}

\end{document}